\documentclass[runningheads]{llncs}
\usepackage{graphicx}

\usepackage{tikz}
\usepackage{comment}
\usepackage{amsmath,amssymb} 
\usepackage{color}
\usepackage{orcidlink}

\usepackage[accsupp]{axessibility}  

\begin{document}
\pagestyle{headings}
\mainmatter
\def\ECCVSubNumber{4073}  

\title{Self-distilled Feature Aggregation for Self-supervised Monocular Depth Estimation} 

\titlerunning{SDFA-Net}
%

\renewcommand{\thefootnote}{\fnsymbol{footnote}}
\author{Zhengming Zhou\inst{1,2}\orcidlink{0000-0002-6792-0739}\index{Zhou, Zhengming} \and
Qiulei Dong\inst{\footnotemark[1] 1,2,3}\orcidlink{0000-0003-4015-1615}\index{Dong, Qiulei}}
\authorrunning{Z. Zhou and Q. Dong}
%
\institute{National Laboratory of Pattern Recognition, Institute of Automation, Chinese Academy of Sciences, Beijing 100190, China \and
School of Artificial Intelligence, University of Chinese Academy of Sciences, Beijing 100049, China \and
Center for Excellence in Brain Science and Intelligence Technology, Chinese Academy of Sciences, Beijing 100190, China \\
\email{zhouzhengming2020@ia.ac.cn}\\
\email{qldong@nlpr.ia.ac.cn}}

\maketitle

\begin{abstract}
Self-supervised monocular depth estimation has received much attention recently in computer vision.
Most of the existing works in literature aggregate multi-scale features for depth prediction via either straightforward concatenation or element-wise addition, however, such feature aggregation operations generally neglect the contextual consistency between multi-scale features.
Addressing this problem, we propose the Self-Distilled Feature Aggregation (SDFA) module for simultaneously aggregating a pair of low-scale and high-scale features and maintaining their contextual consistency.
The SDFA employs three branches to learn three feature offset maps respectively: one offset map for refining the input low-scale feature and the other two for refining the input high-scale feature under a designed self-distillation manner.
Then, we propose an SDFA-based network for self-supervised monocular depth estimation, and design a self-distilled training strategy to train the proposed network with the SDFA module.
Experimental results on the KITTI dataset demonstrate that the proposed method outperforms the comparative state-of-the-art methods in most cases.
The code is available at https://github.com/ZM-Zhou/SDFA-Net\_pytorch.

\end{abstract}

\footnotetext[1]{Corresponding author}

\section{Introduction}
Monocular depth estimation is a challenging topic in computer vision, which aims to predict pixel-wise scene depths from single images.
Recently, self-supervised methods~\cite{Garg2016Unsupervised,Godard2017Unsupervised,Gonzalezbello2020Forget} for monocular depth learning have received much attention, due to the fact that they could be trained without ground truth depth labels.

\begin{figure}[t]
  \centering
  \scriptsize
  \includegraphics[width=9.6cm]{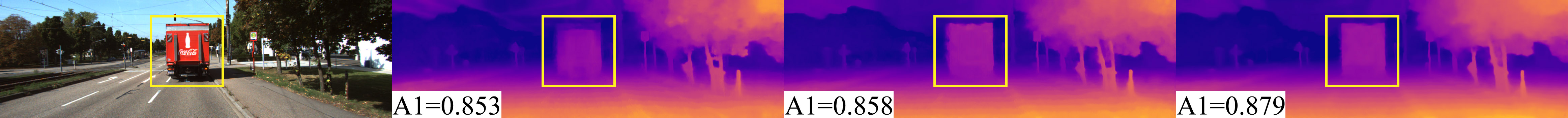}
  \includegraphics[width=9.6cm]{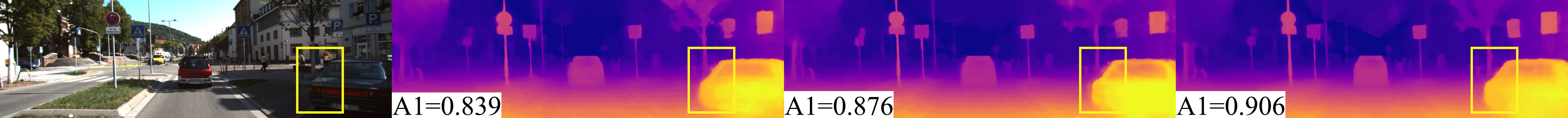}
  \includegraphics[width=9.6cm]{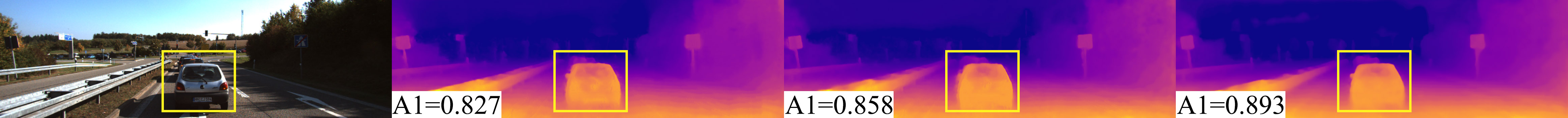}
  \leftline{\qquad\qquad\qquad Input images \qquad\qquad\  Raw \qquad\qquad\qquad\ \  OA \qquad\qquad\quad\  SDFA-Net}
  \caption{
  Depth map comparison by the networks (described in Section~\ref{sec:abla}) with different feature aggregation techniques (the straightforward concatenation (Raw), the Offset-based Aggregation (OA)~\cite{huang2021alignseg}, and the proposed SDFA module) on KITTI~\cite{Geiger2012We}, where `A1' is an accuracy metric.
  }
  \label{fig:head}
\end{figure}

The existing methods for self-supervised monocular depth estimation could be generally categorized into two groups according to the types of training data: the methods which are trained with monocular video sequences~\cite{johnston2020self,wang2021can,Zhou2017Unsupervised} and the methods which are trained with stereo pairs~\cite{Garg2016Unsupervised,Godard2017Unsupervised,Gonzalezbello2020Forget}.
Regardless of the types of training data, many existing methods~\cite{Godard2019Digging,Gonzalezbello2020Forget,peng2021excavating,pilzer2019refine,Shu2020Feature-metric,Watson2019Self,poggi2020uncertainty} employ various encoder-decoder architectures for depth prediction, and the estimation processes could be considered as a general process that sequentially learns multi-scale features and predicts scene depths.
In most of these works, their encoders extract multi-scale features from input images, and their decoders gradually aggregate the extracted multi-scale features via either straightforward concatenation or element-wise addition, however, although such feature aggregation operations have demonstrated their effectiveness to some extent in these existing works, they generally neglect the contextual consistency between the multi-scale features, i.e., the corresponding regions of the features from different scales should contain the contextual information for similar scenes.
This problem might harm a further performance improvement of these works.

Addressing this problem, we firstly propose the Self-Distilled Feature Aggregation (SDFA) module for simultaneously aggregating a pair of low-scale and high-scale features and maintaining their contextual consistency, inspired by the success of a so-called `feature-alignment' module that uses an offset-based aggregation technique for handling the image segmentation task~\cite{huang2021fapn,huang2021alignseg,li2020semantic,li2020improving,cardace2022shallow}.
It has to be pointed out that such a feature-alignment module could not guarantee its superior effectiveness in the self-supervised monocular depth estimation community.
As shown in Figure~\ref{fig:head}, when the Offset-based Aggregation (OA) technique is straightforwardly embedded into an encoder-decoder network for handling the monocular depth estimation task, although it performs better than the concatenation-based aggregation (Raw) under the metric `A1' which is considered as an accuracy metric, it is prone to assign inaccurate depths into pixels on occlusion regions (e.g., the regions around the contour of the vehicles in the yellow boxes in Figure~\ref{fig:head}) due to the calculated inaccurate feature alignment by the feature-alignment module for these pixels. 
In order to solve this problem, the proposed SDFA module employs three branches to learn three feature offset maps respectively: one offset map is used for refining the input low-scale feature, and the other two are jointly used for refining the input high-scale feature under a designed self-distillation manner.

Then, we propose an SDFA-based network for self-supervised monocular depth estimation, called SDFA-Net, which is trained with a set of stereo image pairs.
The SDFA-Net employs an encoder-decoder architecture, which uses a modified version of tiny Swin-transformer~\cite{liu2021swin} as its encoder and an ensemble of multiple SDFA modules as its decoder.
In addition, a self-distilled training strategy is explored to train the SDFA-Net, which selects reliable depths by two principles from a raw depth prediction, and uses the selected depths to train the network by self-distillation.

In sum, our main contributions include:
\begin{itemize}
  \item We propose the Self-Distilled Feature Aggregation (SDFA) module, which could effectively aggregate a low-scale feature with a high-scale feature under a self-distillation manner.
  \item We propose the SDFA-Net with the explored SDFA module for self-supervised monocular depth estimation, where an ensemble of SDFA modules is employed to both aggregate multi-scale features and maintain the contextual consistency among multi-scale features for depth prediction.
  \item We design the self-distilled training strategy for training the proposed network with the explored SDFA module.
  The proposed network achieves a better performance on the KITTI dataset~\cite{Geiger2012We} than the comparative state-of-the-art methods in most cases as demonstrated in Section~\ref{sec:experiments}.
\end{itemize}

\section{Related Work}
Here, we review the two groups of self-supervised monocular depth estimation methods, which are trained with monocular videos and stereo pairs respectively.
\subsection{Self-supervised Monocular Training}
The methods trained with monocular video sequences aim to simultaneously estimate the camera poses and predict the scene depths. 
An end-to-end method was proposed by Zhou et al.~\cite{Zhou2017Unsupervised}, which comprised two separate networks for predicting the depths and camera poses.
Guizilini et al.~\cite{Guizilini20203d} proposed PackNet where the up-sampling and down-sampling operations were re-implemented by 3D convolutions.
Godard et al.~\cite{Godard2019Digging} designed the per-pixel minimum reprojection loss, the auto-mask loss, and the full-resolution sampling in Monodepth2.
Shu et al.~\cite{Shu2020Feature-metric} designed the feature-metric loss defined on feature maps for handling less discriminative regions in the images.

Additionally, the frameworks which learnt depths by jointly using monocular videos and extra semantic information were investigated in~\cite{cheng2020s,Guizilini2020Semantically-guided,jung2021fine,klingner2020self}.
Some works~\cite{Chen2019Self,jiao2021effiscene,Yin2018Geonet} investigated jointly learning the optical flow, depth, and camera pose.
Some methods were designed to handle self-supervised monocular depth estimation in the challenging environments, such as the indoor environments~\cite{ji2021monoindoor,zhou2019moving} and the nighttime environments~\cite{liu2021self,wang2021regularizing}.

\subsection{Self-supervised Stereo Training}
The methods trained with stereo image pairs generally estimate scene depths by predicting the disparity between an input pair of stereo images.
A pioneering work was proposed by Garg et al.~\cite{Garg2016Unsupervised}, which used the predicted disparities and one image of a stereo pair to synthesize the other image at the training stage.
Godard et al.~\cite{Godard2017Unsupervised} proposed a left-right disparity consistency loss to improve the robustness of monocular depth estimation.
Tosi et al.~\cite{Tosi2019Learning} proposed monoResMatch which employed three hourglass-structure networks for extracting features, predicting raw disparities and refining the disparities respectively.
FAL-Net~\cite{Gonzalezbello2020Forget} was proposed to learn depths under an indirect way, where the disparity was represented by the weighted sum of a set of discrete disparities and the network predicted the probability map of each discrete disparity. 
Gonzalez and Kim~\cite{bello2021self} proposed the ambiguity boosting, which improved the accuracy and consistency of depth predictions.

Additionally, for further improving the performance of self-supervised monocular depth estimation, several methods used some extra information (e.g., disparities generated with either traditional algorithms~\cite{Tosi2019Learning,Watson2019Self,Zhu2020The} or extra networks~\cite{chen2021revealing,choi2021adaptive,guo2018learning}, and semantic segmentation labels~\cite{Zhu2020The}).
For example, Watson et al.~\cite{Watson2019Self} proposed depth hints which calculated with Semi Global Matching~\cite{Hirschmuller2005Accurate} and used to guide the network to learn accurate depths.
Other methods employed knowledge distillation~\cite{gou2021knowledge} for self-supervised depth estimation~\cite{pilzer2019refine,peng2021excavating}.
Peng et al.~\cite{peng2021excavating} generated an optimal depth map from the multi-scale outputs of a network, and trained the same network with this distilled depth map.

\section{Methodology}
In this section, we firstly introduce the architecture of the proposed network.
Then, the Self-Distilled Feature Aggregation (SDFA) is described in detail.
Finally, the designed self-distilled training strategy is given.

\subsection{Network Architecture}
\begin{figure}[t]
  \centering
  \includegraphics[width=11.5cm]{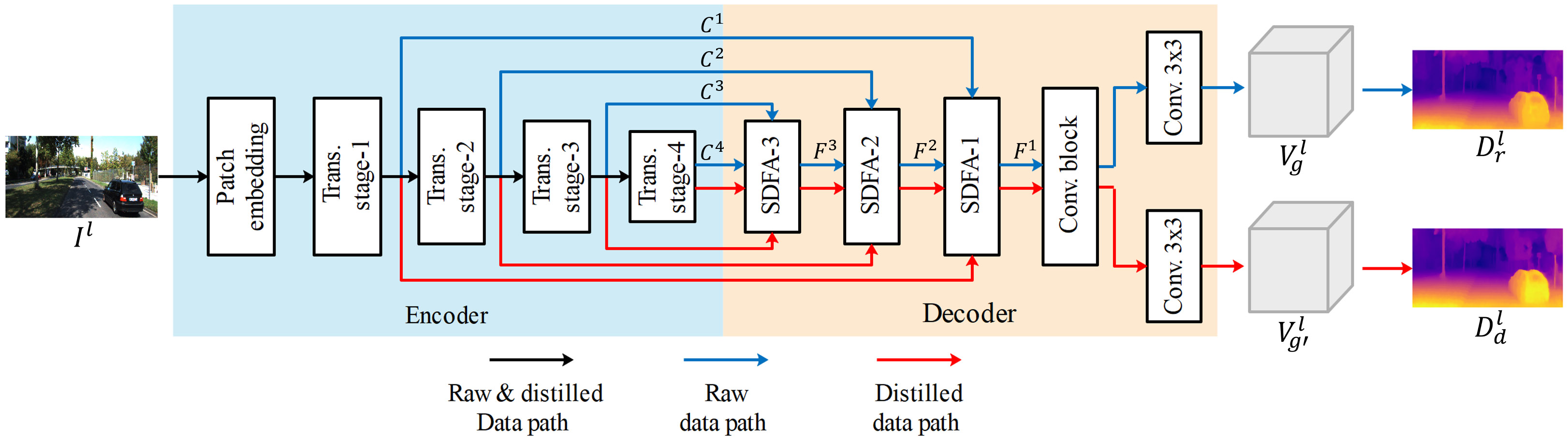}
  \caption{Architecture of SDFA-Net.
           SDFA-Net is used to predict the volume-based depth representation $V_g^l$ and the depth map $D^l$ from the input image $I^l$.
           The features extracted by the encoder could be passed through the raw and distilled data paths to predict raw and distilled depth maps $D^l_{r}, D^l_{d}$ respectively.}
  \label{fig:arch}
\end{figure}

The proposed SDFA-Net adopts an encoder-decoder architecture with skip connections for self-supervised monocular depth estimation, as shown in Figure~\ref{fig:arch}.

\noindent\textbf{Encoder.}
Inspired by the success of vision transformers~\cite{cheng2021swin,dosovitskiy2020vit,ranftl2021vision,wang2021pyramid,yang2021transformer} in various visual tasks, we introduce the following modified version of tiny Swin-transformer~\cite{liu2021swin} as the backbone encoder to extract multi-scale features from an input image $I^l \in \mathbb{R}^{3\times H \times W}$.
The original Swin-transformer contains four transformer stages, and it pre-processes the input image through a convolutional layer with stride=4, resulting in 4 intermediate features $[C_1, C_2,$ $ C_3, C_4]$ with the resolutions of $[\frac{H}{4} \times \frac{W}{4}, \frac{H}{8} \times \frac{W}{8}, \frac{H}{16} \times \frac{W}{16}, \frac{H}{32} \times \frac{W}{32}]$.
Considering rich spatial information is important for depth estimation, we change the convolutional layer with stride=4 in the original Swin-transformer to that with stride=2 in order to keep more high-resolution image information, and accordingly, the resolutions of the features extracted from the modified version of Swin-transformer are twice as they were, i.e. $[\frac{H}{2} \times \frac{W}{2}, \frac{H}{4} \times \frac{W}{4}, \frac{H}{8} \times \frac{W}{8}, \frac{H}{16} \times \frac{W}{16}]$.

\noindent\textbf{Decoder.}
The decoder uses the multi-scale features $\{C^i\}_{i=1}^4$ extracted from the encoder as its input, and it outputs the disparity-logit volume $V^l_g$ as the scene depth representation.
The decoder is comprised of three SDFA modules (denoted as SDFA-1, SDFA-2, SDFA-3 as shown in Figure~\ref{fig:arch}), a convolutional block and two convolutional output layers.
The SDFA module is proposed for adaptively aggregating the multi-scale features with learnable offset maps, which would be described in detail in Section~\ref{sec:SDFA}.
The convolutional block is used for restoring the spatial resolution of the aggregated feature to the size of the input image, consisting of a nearest up-sampling operation and two $3 \times 3$ convolutional layers with the ELU activation~\cite{clevert2015fast}.
For training the SDFA modules under the self-distillation manner and avoiding training two networks, the encoded features could be passed through the decoder via two data paths, and be aggregated by different offset maps.
The two $3 \times 3$ convolutional output layers are used to predict two depth representations at the training stage, which are defined as the raw and distilled depth representations respectively.
Accordingly, the two data paths are defined as the raw data path and the distilled data path.
Once the proposed network is trained, given an arbitrary test image, only the outputted distilled depth representation $V^l_{g'}$ is used as its final depth prediction at the inference stage.

\subsection{Self-Distilled Feature Aggregation}
\label{sec:SDFA}
\begin{figure}[t]
  \centering
  \includegraphics[width=9cm]{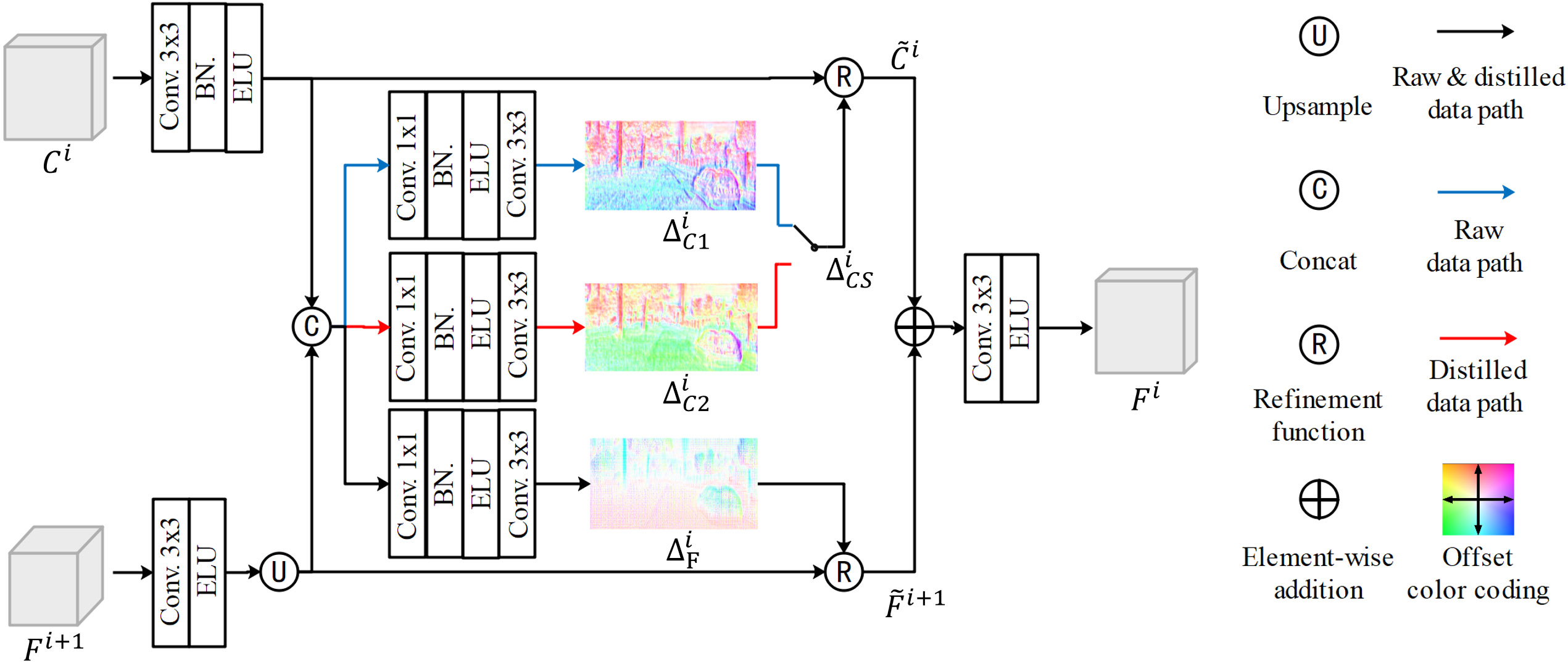}
  \caption{Self-Distilled Feature Aggregation.
           Different offset maps are chosen in different data paths.}
  \label{fig:sdfa}
\end{figure}

The Self-Distilled Feature Aggregation (SDFA) module with learnable feature offset maps is proposed to adaptively aggregate the multi-scale features and maintain the contextual consistency between them.
It jointly uses a low-scale decoded feature from the previous layer and its corresponding feature from the encoder as the input, and it outputs the aggregated feature as shown in Figure~\ref{fig:sdfa}.
The two features are inputted from either raw or distilled data path.

As seen from Figure~\ref{fig:sdfa}, under the designed module SDFA-$i$ ($i=1,2,3$), the feature $F^{i+1}$ (specially $F^4 = C^4$) from its previous layer is passed through a $3\times 3$ convolutional layer with the ELU activation~\cite{clevert2015fast} and up-sampled by the standard bilinear interpolation.
Meanwhile, an additional $3\times 3$ convolutional layer with the Batch Normalization (BN)~\cite{ioffe2015batch} and ELU activation is used to adjust the channel dimensions of the corresponding feature $C^{i}$ from the encoder to be the same as that of $F^{i+1}$.
Then, the obtained two features via the above operations are concatenated together and passed through three branches for predicting the offset maps $\Delta_{F}^i$, $\Delta_{C1}^i$, and $\Delta_{C2}^i$.
The offset map $\Delta_{F}^i$ is used to refine the up-sampled $F^{i+1}$.
According to the used data path $P$, a switch operation `$ \mathcal{S}(\Delta_{C1}^i, \Delta_{C2}^i| P)$' is used to select an offset map from $\Delta_{C1}^i$ and $\Delta_{C2}^i$ for refining the adjusted $C^{i}$, which is formulated as:
\begin{equation}
  \Delta_{CS}^i = \mathcal{S}(\Delta_{C1}^i, \Delta_{C2}^i| P) = 
  \left\{\begin{matrix}
    \Delta_{C1}^i ,&\quad P={\rm raw}  \\
    \Delta_{C2}^i ,& \quad P={\rm distilled} \\
   \end{matrix}\right. \quad. 
\end{equation}

After obtaining the offset maps, a refinement function `$\mathcal{R}(F, \Delta)$' is designed to refine the feature $F$ by the guidance of the offset map $\Delta$, which is implemented by the bilinear interpolation kernel.
Specifically, to generate a refined feature $\tilde{F}(p)$ on the position $p=[x, y]^\top$ from $F\in\mathbb{R}^{C\times H \times W}$ with an offset $\Delta(p)$, the refinement function is formulated as:
\begin{equation}
  \begin{split}
    \tilde{F}(p) = \mathcal{R}(F, \Delta(p))
    = \left< F(p+\Delta(p)) \right>\quad,
  \end{split}
\end{equation}
where `$\left<\cdot\right>$' denotes the bilinear sampling operation.
Accordingly, the refined features $\tilde{F}^{i+1}$ and $\tilde{C}^i$ are generated with:
\begin{equation}
  \tilde{F}^{i+1} = \mathcal{R}
    \left(\mathcal{U}
      \left({\rm conv}
        \left(F^{i+1}
        \right)
      \right), \Delta_{F}^i
    \right) \quad,
\end{equation}
\begin{equation}
  \tilde{C}^{i} = \mathcal{R}
    \left({\rm convb}
      \left(C^{i}
      \right), \Delta_{CS}^i
    \right) \quad,
\end{equation}
where `$\mathcal{U}(\cdot)$' denotes the bilinear up-sample, `$\rm conv(\cdot)$' denotes the $3\times 3$ convolutional layer with the ELU activation, and `$\rm convb(\cdot)$' denotes the $3\times 3$ convolutional layer with the BN and ELU.
Finally, the aggregated feature $F^i$ is obtained with the two refined features as:
\begin{equation}
  F^{i} = {\rm conv}\left(\tilde{C}^{i} \oplus \tilde{F}^{i+1}\right) \quad,
\end{equation}
where `$\oplus$' denotes the element-wise addition.

Considering that (i) the offsets learnt with self-supervised training are generally suboptimal because of occlusions and image ambiguities and (ii) more effective offsets are expected to be learnt by utilizing extra clues  (e.g., some reliable depths from the predicted depth map in Section~\ref{sec:train_stra}), SDFA is trained in the following designed self-distillation manner.

\subsection{Self-distilled Training Strategy}
\label{sec:train_stra}
\begin{figure}[t]
  \centering
  \includegraphics[width=11.5cm]{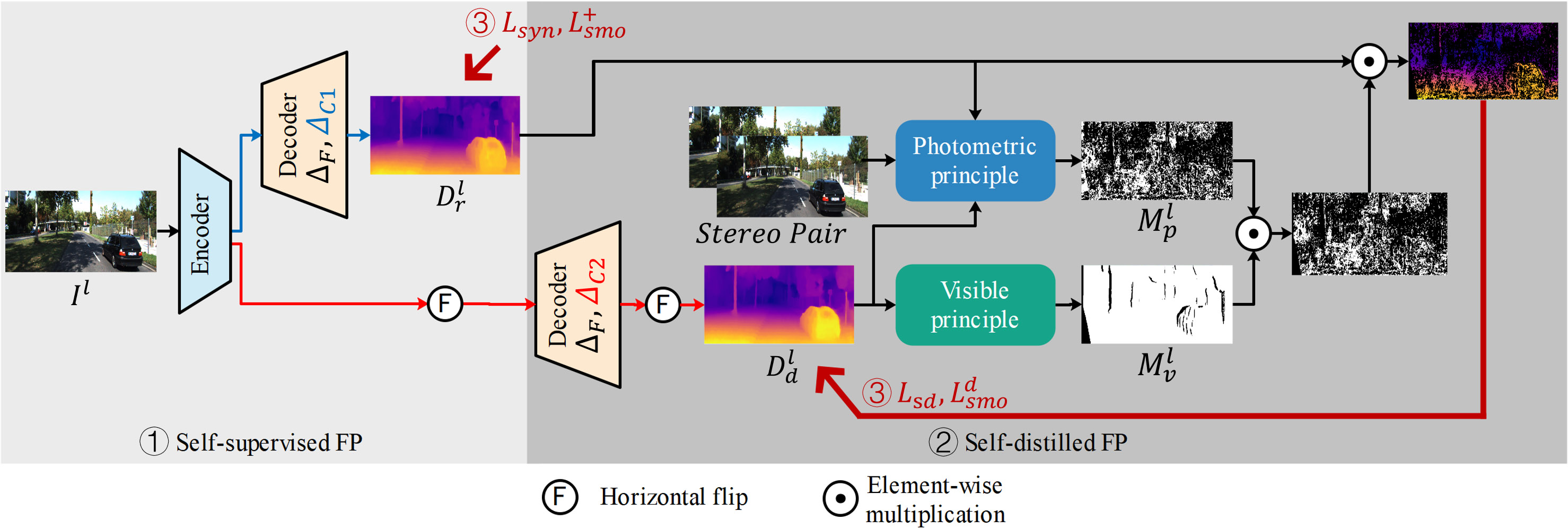}
  \caption{Self-distilled training strategy.
           It comprises three steps: the self-supervised Forward Propagation (FP), the self-distilled Forward Propagation (FP), and the loss computation.}
  \label{fig:distill}
\end{figure}

In order to train the proposed network with a set of SDFA modules, we design a self-distilled training strategy as shown in Figure~\ref{fig:distill}, which divides each training iteration into three sequential steps: the self-supervised forward propagation, the self-distilled forward propagation and the loss computation.

\noindent\textbf{Self-supervised Forward Propagation.}
In this step, the network takes the left image $I^l$ in a stereo pair as the input, outputting a left-disparity-logit volume $V_{g}^l \in \mathbb{R}^{N\times H\times W}$ via the raw data path. 
As done in \cite{Gonzalezbello2020Forget}, $I^l$ and $V_{g}^l$ are used to synthesize the right image $\hat{I}^r$, while the raw depth map $D_{r}^l$ is obtained with $V_{g}^l$. 

Specifically, given the minimum and maximum disparities $d_{\rm min}$ and $d_{\rm max}$, we firstly define the discrete disparity level $d_n$ by the exponential quantization~\cite{Gonzalezbello2020Forget}:
\begin{equation}
  d_n = d_{\rm max} \left(\frac{d_{\rm min}}{d_{\rm max}}\right)^{\frac{n}{N-1}}, \quad n=0, 1, ..., N-1 \quad,
\end{equation}
where $N$ is the total number of the disparity levels.
For synthesizing the right images, each channel of $V_{g}^l$ is shifted with the corresponding disparity $d_n$, generating the right-disparity-logit volume $\hat{V}_{g}^r$.
A right-disparity-probability volume $\hat{V}_{p}^r$ is obtained by passing $\hat{V}_{g}^r$ through the softmax operation and is employed for stereoscopic image synthesis as:
\begin{equation}
  \hat{I}^r=\sum_{n=0}^{N-1} \hat{V}_{p|n}^r \odot I^l_n \quad,
\end{equation}
where $\hat{V}_{p|n}^r$ is the $n^{\rm th}$ channel of $\hat{V}_{p}^r$, `$\odot$' denotes the element-wise multiplication, and $I^l_n$ is the left image shifted with $d_n$.
For obtaining the raw depth map, $V_{g}^l$ is passed through the softmax operation to generate the left-disparity-probability volume $V_{p}^l$. 
According to the stereoscopic image synthesis, the $n^{\rm th}$ channel of $V_{p}^l$ approximately equals the probability map of $d_n$.
And a pseudo-disparity map $d^l_{+}$ is obtained as:
\begin{equation}
  d^l_{+} = \sum_{n=0}^{N-1}V_{p|n}^l \cdot d_n \quad.
  \label{equ:volume2d}
\end{equation}
Given the baseline length $B$ of the stereo pair and the horizontal focal length $f_x$ of the left camera, the raw depth map $D_{r}^l$ for the left image is calculated via $D_{r}^l = \frac{Bf_x}{d_{+}^l}$.

\noindent\textbf{Self-distilled Forward Propagation.} 
For alleviating the influence of occlusions and learning more accurate depths, the multi-scale features $\{C^i\}_{i=1}^4$ extracted by the encoder are horizontally flipped and passed through the decoder via the distilled data path, outputting a new left-disparity-logit volume $V_{g'}^l$.
After flipping $V_{g'}^l$ back, the distilled disparity map $d^l_{d}$ and distilled depth map $D^l_{d}$ are obtained by the flipped $V_{g'}^l$ as done in Equation (\ref{equ:volume2d}).
For training the SDFA-Net under the self-distillation manner, we employs two principles to select reliable depths from the raw depth map $D_{r}^l$ : the photometric principle and the visible principle.
As shown in Figure~\ref{fig:distill}, we implement the two principles with two binary masks, respectively.

The photometric principle is used to select the depths from $D^l_{r}$ in pixels where they make for a superior reprojected left image.
Specifically, for an arbitrary pixel coordinate $p$ in the left image, its corresponding coordinate $p'$ in the right image is obtained with a left depth map $D^l$ as:
\begin{equation}
    p'=p-\left[\frac{Bf_x}{D^l(p)},0\right]^\top
    \quad.
    \label{equ:corr}
\end{equation}
Accordingly, the reprojected left image $\hat{I}^l$ is obtained by assigning the RGB value of the right image pixel $p'$ to the pixel $p$ of $\hat{I}^l$.
Then, a weighted sum of the $L_1$ loss and the structural similarity (SSIM) loss ~\cite{Wang2004Image}
is employed to measure the photometric difference between the reprojected image $\hat{I}^l$ and raw image $I^l$:
\begin{equation}
    l(D^l) = \left(\alpha \left \| \hat{I}^l -  I^l \right \|_1
    +(1 - \alpha) {\rm SSIM}{(\hat{I}^l, I^l)} \right) \quad,
\end{equation}
where  $\alpha$ is a balance parameter.
The photometric principle is formulated as:
\begin{equation}
    M^l_{p} = \left[\left(l\left(D^l_{r}\right) - l\left(D^l_{d}\right) < \epsilon\right) \cap l\left(D^l_{r}\right) < t_1 \right] \quad,
\end{equation}
where `$[\cdot]$' is the iverson bracket, $\epsilon$ and $t_1$ are predefined thresholds.
The second term in the iverson bracket is used to ignore the inaccurate depths with large photometric errors.

The visible principle is used to eliminate the potential inaccurate depths in the regions that are visible only in the left image, e.g., the pixels occluded in the right image or out of the edge of the right image.
The pixels which are occluded in the right image could be found with the left disparity map $d^l$.
Specifically, it is observed that for an arbitrary pixel location $p = [x,y]^\top$ and its horizontal right neighbor $p_i = [x+i,y]^\top (i=1,2,...,K)$ in the left image, if the corresponding location of $p$ is occluded by that of $p_i$ in the right image, the difference between their disparities should be close or equal to the difference of their horizontal coordinates~\cite{Zhu2020The}.
Accordingly, the occluded mask calculated with the distilled disparity map $d^l_d$ is formulated as:
\begin{equation}
  M^l_{occ}(p)=\left[
      \min_{i}
      \left(
        \left|d^l_{d}(p_i) - d^l_{d}(p) - i \right|
      \right) < t_2\right] \quad,
\end{equation}
where $t_2$ is a predefined threshold.
Additionally, an out-of-edge mask $M^l_{out}$~\cite{Mahjourian2018Unsupervised} is jointly utilized for filtering out the pixels whose corresponding locations are out of the right image.
Accordingly, the visible principle is formulated as:
\begin{equation}
  M^l_{v}=M^l_{occ} \odot M^l_{out}\quad.
\end{equation}

\noindent\textbf{Loss Computation.} 
In this step, the total loss is calculated for training the SDFA-Net.
It is noted that the raw depth map $D^l_{r}$ is learnt by maximizing the similarity between the real image $I^r$ and the synthesized image $\hat{I}^r$, while the distilled depth map $D^l_{d}$ is learnt with self-distillation.
The total loss function is comprised of four terms: the image synthesis loss $L_{syn}$, the self-distilled loss $L_{sd}$, the raw smoothness loss $L_{smo}^{+}$, and the distilled smoothness loss $L_{smo}^{d}$.

The image synthesis loss $L_{syn}$ contains the $L_1$ loss and the perceptual loss \cite{Johnson2016Perceptual} for reflecting the similarity between the real right image $I^r$ and the synthesized right image $\hat{I}^r$:
\begin{equation}
  L_{syn}= \left\|\hat{I}^r - I^r \right \|_1 + \beta \sum_{i=1,2,3} \left\|\phi_i (\hat{I}^r) - \phi_i (I^r) \right \|_2
  \quad,
\end{equation}
where `$\| \cdot \|_1$' and `$\| \cdot \|_2$' represent the $L_1$ and $L_2$ norm, $\phi_i(\cdot)$ denotes the output of $i^{\rm th}$ pooling layer of a pretrained VGG19 \cite{simonyan2014very}, and $\beta$ is a balance parameter.

The self-distilled loss $L_{sd}$ adopts the $L_1$ loss to distill the accurate depths from the raw depth map $D^l_{r}$ to the distilled depth map $D^l_{d}$, where the accurate depths are selected by the photometric and visible masks $M^l_{p}$ and $M^l_{v}$:
\begin{equation}
  L_{sd} = M^l_{p} \odot M^l_{v} \odot \left\| D^l_{d} - D^l_{r} \right\|_1 \quad.
\end{equation}

The edge-aware smoothness loss $L_{smo}$ is employed for constraining the continuity of the pseudo disparity map $d^l_{+}$ and the distilled disparity map $d^l_{d}$:
\begin{equation}
  L_{smo}^{+}= \left \| \partial_x d^l_{+} \right \|_1 e^{-\gamma \left \| \partial_x I^l \right \|_1}
  + \left \| \partial_y d^l_{+} \right \|_1 e^{-\gamma \left \| \partial_y I^l \right \|_1}\
  \quad,
\end{equation}
\begin{equation}
  L_{smo}^{d}= \left \| \partial_x d^l_{d} \right \|_1 e^{-\gamma \left \| \partial_x I^l \right \|_1}
  + \left \| \partial_y d^l_{d} \right \|_1 e^{-\gamma \left \| \partial_y I^l \right \|_1}\
  \quad,
\end{equation}
where `$\partial_x$', `$\partial_y$' are the differential operators in the horizontal and vertical directions respectively, and $\gamma$ is a parameter for adjusting the degree of edge preservation.

Accordingly, the total loss is a weighted sum of the above four terms, which is formulated as:
\begin{equation}
  L = L_{syn} + \lambda_1 L_{smo}^{+} + \lambda_2 L_{sd} + \lambda_3 L_{smo}^{d} \quad,
\end{equation}
where $\{\lambda_1, \lambda_2, \lambda_3\}$ are three preseted weight parameters.
Considering that the depths learnt under the self-supervised manner are unreliable at the early training stage, $\lambda_2$ and $\lambda_3$ are set to zeros at these training iterations, while the self-distilled forward propagation is disabled.

\section{Experiments}
\label{sec:experiments}
In this section, we evaluate the SDFA-Net as well as 15 state-of-the-art methods and perform ablation studies on the KITTI dataset~\cite{Geiger2012We}.
The Eigen split~\cite{Eigen2014Depth} of KITTI which comprises 22600 stereo pairs for training and 679 images for testing is used for network training and testing, while the improved Eigen test set~\cite{Uhrig2017Sparsity} is also employed for network testing.
Additionally, 22972 stereo pairs from the Cityscapes dataset~\cite{Cordts2016The} are jointly used for training as done in~\cite{Gonzalezbello2020Forget}.
At both the training and inference stages, the images are resized into the resolution of $1280 \times 384$.
We utilize the crop proposed in~\cite{Garg2016Unsupervised} and the standard depth cap of 80m in the evaluation.
The following error and accuracy metrics are used as done in the existing works~\cite{Godard2019Digging,Gonzalezbello2020Forget,klingner2020self,peng2021excavating,Shu2020Feature-metric}: Abs Rel, Sq Rel, RMSE, logRMSE, A1~$=\delta < 1.25$, A2~$=\delta < 1.25^2$, and A3~$=\delta < 1.25^3$.
Please see the supplemental material for more details about the datasets and metrics.

\subsection{Implementation Details}
We implement the SDFA-Net with PyTorch~\cite{Paszke2019Pytorch}, and the modified version of tiny Swin-transformer is pretrained on the ImageNet1K dataset~\cite{deng2009imagenet}.
The Adam optimizer~\cite{Kingma2014Adam} with $\beta_1=0.5$ and $\beta_2=0.999$ is used to train the SDFA-Net for 50 epochs with a batch size of 12.
The initial learning rate is firstly set to $10^{-4}$, and is downgraded by half at epoch 30 and 40.
The self-distilled forward propagation and the corresponding losses are used after training 25 epochs.
For calculating disparities from the outputted volumes, we set the minimum and the 
maximum disparities to $d_{\min}=2,d_{\max}=300$, and the number of the discrete levels are set to $N=49$.
The weight parameters for the loss function are set to $\lambda_1=0.0008, \lambda_2=0.01$ and $\lambda_3=0.0016$, while we set $\beta=0.01$ and $\gamma=2$.
For the two principles in the self-distilled forward propagation, we set $\alpha=0.15, \epsilon = 1e-5, t_1=0.2, t_2=0.5$ and $K=61$.
We employ random resizing (from 0.75 to 1.5) and cropping (192$\times$640), random horizontal flipping, and random color augmentation as the data augmentations.

\begin{table}[t]
  \centering
  \caption{Quantitative comparison on both the raw and improved KITTI Eigen test sets.
          $\downarrow/\uparrow$ denotes that lower / higher is better.
          The best and the second best results are in \textbf{bold} and \underline{underlined} in each metric.
          }
  \tiny
  \renewcommand\tabcolsep{1.3pt}
  \begin{tabular}{|lcccc|cccc|ccc|}
  \hline
  Method &
    PP &
    Sup. &
    Data. &
    Resolution &
    Abs Rel $\downarrow$ &
    Sq Rel $\downarrow$ &
    RMSE $\downarrow$ &
    logRMSE $\downarrow$ &
    A1 $\uparrow$ &
    A2 $\uparrow$ &
    A3 $\uparrow$ \\\hline
  \multicolumn{12}{|c|}{Raw Eigen test set} \\\hline
  R-MSFM6~\cite{zhou2021r} &
    &
   M &
   K &
   320x1024 &
   0.108 &
   0.748 &
   4.470 &
   0.185 &
   0.889 &
   0.963 &
   0.982 \\
  PackNet~\cite{Guizilini20203d} &
     &
    M &
    K &
    384x1280 &
    0.107 &
    0.802 &
    4.538 &
    0.186 &
    0.889 &
    0.962 &
    0.981 \\
  SGDepth~\cite{klingner2020self} &
        &
    M(s) &
    K &
    384x1280 &
    0.107 &
    0.768 &
    4.468 &
    0.186 &
    0.891 &
    0.963 &
    0.982 \\
  FeatureNet~\cite{Shu2020Feature-metric} &
     &
    M &
    K &
    320x1024 &
    0.104 &
    0.729 &
    4.481 &
    0.179 &
    0.893 &
    0.965 &
    \underline{0.984} \\
  monoResMatch~\cite{Tosi2019Learning} &
    \checkmark &
    S(d) &
    K &
    384x1280 &
    0.111 &
    0.867 &
    4.714 &
    0.199 &
    0.864 &
    0.954 &
    0.979 \\
  Monodepth2~\cite{Godard2019Digging} &
    \checkmark &
    S &
    K &
    320x1024 &
    0.105 &
    0.822 &
    4.692 &
    0.199 &
    0.876 &
    0.954 &
    0.977 \\
  DepthHints~\cite{Watson2019Self} &
    \checkmark &
    S(d) &
    K &
    320x1024 &
    0.096 &
    0.710 &
    4.393 &
    0.185 &
    0.890 &
    0.962 &
    0.981 \\
  DBoosterNet-e~\cite{bello2021self} &
     &
    S &
    K &
    384x1280 &
    0.095 &
    0.636 &
    4.105 &
    0.178 &
    0.890 &
    0.963 &
    \underline{0.984} \\
  SingleNet~\cite{chen2021revealing} &
    \checkmark &
    S &
    K &
    320x1024 &
    0.094 &
    0.681 &
    4.392 &
    0.185 &
    0.892 &
    0.962 &
    0.981 \\
  FAL-Net~\cite{Gonzalezbello2020Forget} &
    \checkmark &
    S &
    K &
    384x1280 &
    0.093 &
    0.564 &
    3.973 &
    0.174 &
    0.898 &
    \underline{0.967} &
    \textbf{0.985} \\
  Edge-of-depth~\cite{Zhu2020The} &
    \checkmark &
    S(s,d) &
    K &
    320x1024 &
    0.091 &
    0.646 &
    4.244 &
    0.177 &
    0.898 &
    0.966 &
    0.983 \\
  PLADE-Net~\cite{Gonzalez2021Plade} &
    \checkmark &
    S &
    K &
    384x1280 &
    \textbf{0.089} &
    0.590 &
    4.008 &
    0.172 &
    0.900 &
    \underline{0.967} &
    \textbf{0.985} \\
  EPCDepth~\cite{peng2021excavating} &
    \checkmark &
    S(d) &
    K &
    320x1024 &
    0.091 &
    0.646 &
    4.207 &
    0.176 &
    0.901 &
    0.966 &
    0.983 \\
  \emph{SDFA-Net(Ours)} &
     &
    S &
    K &
    384x1280 &
    \underline{0.090} &
    \underline{0.538} &
    \underline{3.896} &
    \underline{0.169} &
    \underline{0.906} &
    \textbf{0.969} &
    \textbf{0.985} \\
  \emph{SDFA-Net(Ours)} &
    \checkmark &
    S &
    K &
    384x1280 &
    \textbf{0.089} &
    \textbf{0.531} &
    \textbf{3.864} &
    \textbf{0.168} &
    \textbf{0.907} &
    \textbf{0.969} &
    \textbf{0.985} \\\hline
  PackNet~\cite{Guizilini20203d} &
     &
    M &
    CS+K &
    384x1280 &
    0.104 &
    0.758 &
    4.386 &
    0.182 &
    0.895 &
    0.964 &
    0.982 \\
  SemanticGuide~\cite{Guizilini2020Semantically-guided} &
     &
    M(s) &
    CS+K &
    384x1280 &
    0.100 &
    0.761 &
    4.270 &
    0.175 &
    0.902 &
    0.965 &
    0.982 \\
  monoResMatch~\cite{Tosi2019Learning} &
    \checkmark &
    S(d) &
    CS+K &
    384x1280 &
    0.096 &
    0.673 &
    4.351 &
    0.184 &
    0.890 &
    0.961 &
    0.981 \\
  DBoosterNet-e~\cite{bello2021self} &
    \checkmark &
    S &
    CS+K &
    384x1280 &
    0.086 &
    0.538 &
    3.852 &
    0.168 &
    0.905 &
    \underline{0.969} &
    \textbf{0.985} \\
  FAL-Net~\cite{Gonzalezbello2020Forget} &
    \checkmark &
    S &
    CS+K &
    384x1280 &
    0.088 &
    0.547 &
    4.004 &
    0.175 &
    0.898 &
    0.966 &
    \underline{0.984} \\
  PLADE-Net~\cite{Gonzalez2021Plade} &
    \checkmark &
    S &
    CS+K &
    384x1280 &
    0.087 &
    0.550 &
    \underline{3.837} &
    \underline{0.167} &
    0.908 &
    \textbf{0.970} &
    \textbf{0.985} \\
  \emph{SDFA-Net(Ours)} &
     &
    S &
    CS+K &
    384x1280 &
    \underline{0.085} &
    \underline{0.531} &
    3.888 &
    \underline{0.167} &
    \underline{0.911} &
    \underline{0.969} &
    \textbf{0.985} \\
  \emph{SDFA-Net(Ours)} &
    \checkmark &
    S &
    CS+K &
    384x1280 &
    \textbf{0.084} &
    \textbf{0.523} &
    \textbf{3.833} &
    \textbf{0.166} &
    \textbf{0.913} &
    \textbf{0.970} &
    \textbf{0.985} \\\hline
  \multicolumn{12}{|c|}{Improved Eigen test set} \\\hline
  DepthHints~\cite{Watson2019Self} &
    \checkmark &
    S(d) &
    K &
    320x1024 &
    0.074 &
    0.364 &
    3.202 &
    0.114 &
    0.936 &
    0.989 &
    0.997 \\
  WaveletMonoDepth~\cite{ramamonjisoa2021single} &
    \checkmark &
    S(d) &
    K &
    320x1024 &
    0.074 &
    0.357 &
    3.170 &
    0.114 &
    0.936 &
    0.989 &
    0.997 \\
  DBoosterNet-e~\cite{bello2021self} &
     &
    S &
    K &
    384x1280 &
    \underline{0.070} &
    0.298 &
    2.916 &
    0.109 &
    0.940 &
    0.991 &
    \underline{0.998} \\
  FAL-Net~\cite{Gonzalezbello2020Forget} &
    \checkmark &
    S &
    K &
    384x1280 &
    0.071 &
    0.281 &
    \underline{2.912} &
    0.108 &
    0.943 &
    0.991 &
    \underline{0.998} \\
  PLADE-Net~\cite{Gonzalez2021Plade} &
    \checkmark &
    S &
    K &
    384x1280 &
    \textbf{0.066} &
    \underline{0.272} &
    2.918 &
    \underline{0.104} &
    \underline{0.945} &
    \underline{0.992} &
    \underline{0.998} \\
    \emph{SDFA-Net(Ours)} &
    \checkmark &
    S &
    K &
    384x1280 &
    0.074 &
    \textbf{0.228} &
    \textbf{2.547} &
    \textbf{0.101} &
    \textbf{0.956} &
    \textbf{0.995} &
    \textbf{0.999} \\\hline
  PackNet~\cite{Guizilini20203d} &
     &
    M &
    CS+K &
    384x1280 &
    0.071 &
    0.359 &
    3.153 &
    0.109 &
    0.944 &
    0.990 &
    0.997 \\
  DBoosterNet-e~\cite{bello2021self} &
    \checkmark &
    S &
    CS+K &
    384x1280 &
    \textbf{0.062} &
    \underline{0.242} &
    \underline{2.617} &
    \textbf{0.096} &
    \underline{0.955} &
    \underline{0.994} &
    \underline{0.998} \\
  FAL-Net~\cite{Gonzalezbello2020Forget} &
    \checkmark &
    S &
    CS+K &
    384x1280 &
    0.068 &
    0.276 &
    2.906 &
    0.106 &
    0.944 &
    0.991 &
    \underline{0.998} \\
  PLADE-Net~\cite{Gonzalez2021Plade} &
    \checkmark &
    S &
    CS+K &
    384x1280 &
    \underline{0.065} &
    0.253 &
    2.710 &
    \underline{0.100} &
    0.950 &
    0.992 &
    \underline{0.998} \\
    \emph{SDFA-Net(Ours)} &
    \checkmark &
    S &
    CS+K &
    384x1280 &
    0.069 &
    \textbf{0.207} &
    \textbf{2.472} &
    \textbf{0.096} &
    \textbf{0.963} &
    \textbf{0.995} &
    \textbf{0.999}\\\hline
  \end{tabular}
  \label{tab:main}
\end{table}

\subsection{Comparative Evaluation}
We firstly evaluate the SDFA-Net on the raw KITTI Eigen test set~\cite{Eigen2014Depth} in comparison to 15 existing methods listed in Table~\ref{tab:main}.
The comparative methods are trained with either monocular video sequences (M)~\cite{Guizilini20203d,Guizilini2020Semantically-guided,klingner2020self,Shu2020Feature-metric,zhou2021r} or stereo image pairs (S)~\cite{chen2021revealing,Godard2019Digging,Gonzalezbello2020Forget,Gonzalez2021Plade,bello2021self,peng2021excavating,ramamonjisoa2021single,Tosi2019Learning,Watson2019Self,Zhu2020The}.
It is noted that some of them are trained with additional information, such as the semantic segmentation label (s)~\cite{Guizilini2020Semantically-guided,klingner2020self,Zhu2020The}, and the offline computed disparity (d)~\cite{peng2021excavating,ramamonjisoa2021single,Tosi2019Learning,Watson2019Self,Zhu2020The}.
Additionally, we also evaluate the SDFA-Net on the improved KITTI Eigen test set~\cite{Uhrig2017Sparsity}.
The corresponding results are reported in Table~\ref{tab:main}.

As seen from Lines 1-23 of Table~\ref{tab:main}, when only the KITTI dataset~\cite{Geiger2012We} is used for training (K), SDFA-Net performs best among all the comparative methods in most cases even without the post processing (PP.), and its performance is further improved when a post-processing operation step~\cite{Godard2017Unsupervised} is implemented on the raw Eigen test set. 
When both Cityscapes~\cite{Cordts2016The} and KITTI~\cite{Geiger2012We} are jointly used for training (CS+K), the performance of SDFA-Net is further improved, and it still performs best among all the comparative methods on the raw Eigen test set, especially under the metric `Sq. Rel'.

As seen from Lines 24-34 of Table~\ref{tab:main}, when the improved Eigen test set is used for testing, the SDFA-Net still performs superior to the comparative methods in most cases.
Although it performs relatively poorer under the metric `Abs Rel.', its performances under the other metrics are significantly improved.
These results demonstrate that the SDFA-Net could predict more accurate depths.

\begin{figure}[t]
  \centering
  \includegraphics[width=12cm]{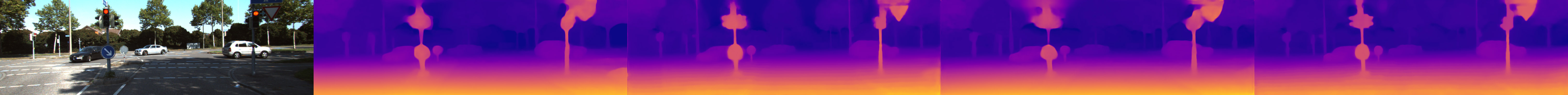}
  \includegraphics[width=12cm]{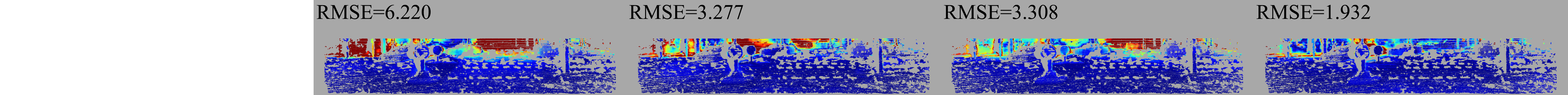}
  \includegraphics[width=12cm]{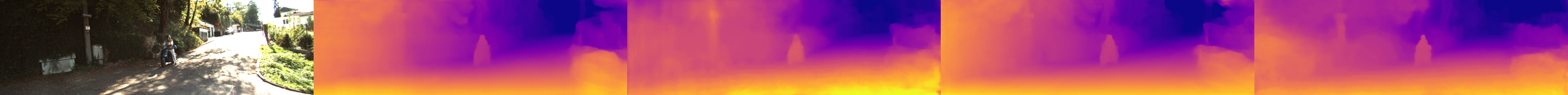}
  \includegraphics[width=12cm]{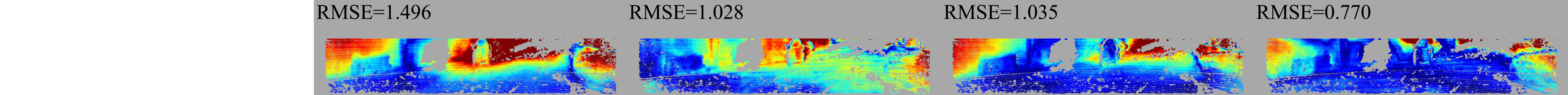}
  \includegraphics[width=12cm]{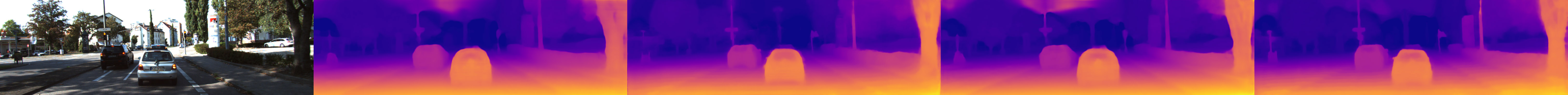}
  \includegraphics[width=12cm]{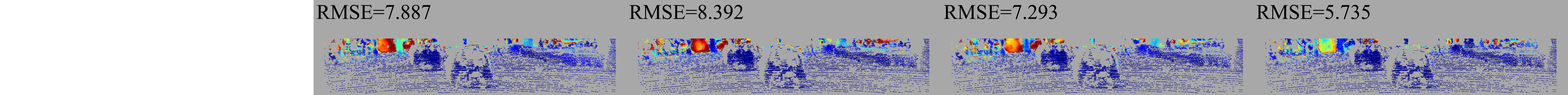}
  \scriptsize
  \leftline{\quad\quad Input Images \quad Edge-of-Depth~\cite{Zhu2020The} \quad FAL-Net~\cite{Gonzalezbello2020Forget} \quad\quad EPCDepth~\cite{peng2021excavating} \quad\quad SDFA-Net(Ours)}
  \caption{Visualization results of Edge-of-Depth~\cite{Zhu2020The}, FAL-Net~\cite{Gonzalezbello2020Forget}, EPCDepth~\cite{peng2021excavating} and our SDFA-Net on KITTI.
  The input images and predicted depth maps are shown in the odd rows, while we give the corresponding `RMSE' error maps calculated with the improved Eigen test set in the even rows.
  For the error maps, red indicates larger error, and blue indicates smaller error.
  }
  \label{fig:main}
\end{figure}

In Figure~\ref{fig:main}, we give several visualization results of SDFA-Net as well as three typical methods whose codes have been released publicly: Edge-of-Depth~\cite{Zhu2020The}, FAL-Net~\cite{Gonzalezbello2020Forget}, and EPCDepth~\cite{peng2021excavating}.
It can be seen that SDFA-Net does not only maintain delicate geometric details, but also predict more accurate depths in distant scene regions as shown in the error maps.

\subsection{Ablation Studies} 
\label{sec:abla}
To verify the effectiveness of each element in SDFA-Net, we conduct ablation studies on the KITTI dataset~\cite{Geiger2012We}.
We firstly train a baseline model only with the self-supervised forward propagation and the self-supervised losses (i.e. the image synthesis loss $L_{syn}$ and the raw smoothness loss $L_{smo}^{+}$).
The baseline model comprises an original Swin-transformer backbone (Swin) as the encoder and a raw decoder proposed in~\cite{Godard2019Digging}.
It is noted that in each block of the raw decoder, the multi-scale features are aggregated by the straightforward concatenation.
Then we replace the backbone by the modified version of the Swin-transformer (Swin$^\dagger$).
We also sequentially replace the $i^{\rm th}$ raw decoder block with the SDFA module, and our full model (SDFA-Net) employs three SDFA modules.
The proposed self-distilled training strategy is used to train the models that contain the SDFA module(s).
Moreover, an Offset-based Aggregation (OA) module which comprises only two offset branches and does not contain the `Distilled data path' is employed for comparing to the SDFA module.
We train the model which uses the OA modules in the decoder with the self-supervised losses only.
The corresponding results are reported in Table~\ref{tab:abla}(a).

\begin{table}[t]
  \centering
  \caption{(a) Quantitative comparison on the raw Eigen test set in the ablation study.
           (b) The average norms (in pixels) of the offset vectors on the raw Eigen test set.}
  \begin{minipage}{8.5cm}
    \centering
    \tiny
    \renewcommand\tabcolsep{1.3pt}
    \begin{tabular}{|l|cccc|ccc|}
      \hline
      \multicolumn{1}{|c|}{Method}         & Abs Rel$\downarrow$ & Sq Rel$\downarrow$ & RMSE$\downarrow$  & logRMSE$\downarrow$ & A1$\uparrow$    & A2$\uparrow$    & A3$\uparrow$    \\\hline
      Swin+Raw                           & 0.102   & 0.615  & 4.323 & 0.184   & 0.887 & 0.963 & 0.983 \\
      Swin$^\dagger$+Raw        & 0.100   & 0.595  & 4.173 & 0.180   & 0.893 & 0.966 & 0.984 \\
      Swin$^\dagger$+SDFA-1 & 0.097   & 0.568  & 3.993 & 0.175   & 0.898 & 0.968 & 0.985 \\
      Swin$^\dagger$+SDFA-2 & 0.099   & 0.563  & 3.982 & 0.174   & 0.900 & 0.968 & 0.985 \\
      Swin$^\dagger$+SDFA-3 & 0.094   & 0.553  & 3.974 & 0.172   & 0.896 & 0.966 & 0.985 \\
      Swin$^\dagger$+OA         & 0.091   & 0.554  & 4.082 & 0.174   & 0.899 & 0.967 & 0.984 \\
      SDFA-Net              & \textbf{0.090}   & \textbf{0.538}  & \textbf{3.896} & \textbf{0.169}   & \textbf{0.906} & \textbf{0.969} & \textbf{0.985} \\\hline
    \end{tabular}
    \scriptsize
    \centerline{(a)}
  \end{minipage}
  \begin{minipage}{3.5cm}
    \centering
    \tiny
    \renewcommand\tabcolsep{1.3pt}
    \begin{tabular}{|c|ccc|}
      \hline
      SDFA-i & $\Delta_F^{i}$ & $\Delta_{C1}^{i}$    & $\Delta_{C2}^{i}$    \\\hline
      1 & 1.39  & 0.14 & 0.59 \\
      2 & 5.85  & 0.53 & 0.56 \\
      3 & 14.89 & 0.65 & 0.65  \\\hline
    \end{tabular}
    \scriptsize
    \centerline{(b)}
  \end{minipage}
  
  \label{tab:abla}
\end{table}

\begin{figure}[t]
  \centering
  \scriptsize
  \includegraphics[width=10cm]{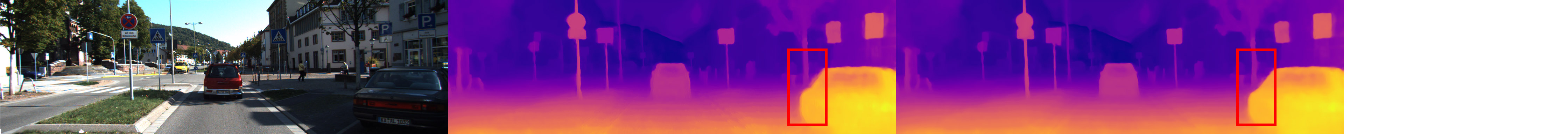}
  \leftline{ \qquad\qquad\quad\  Input image \qquad\qquad\qquad\quad\  $D^{l}_r$\qquad\qquad\qquad\qquad\quad\   $D^{l}_d$}
  \centerline{(a)}
  \includegraphics[width=10cm]{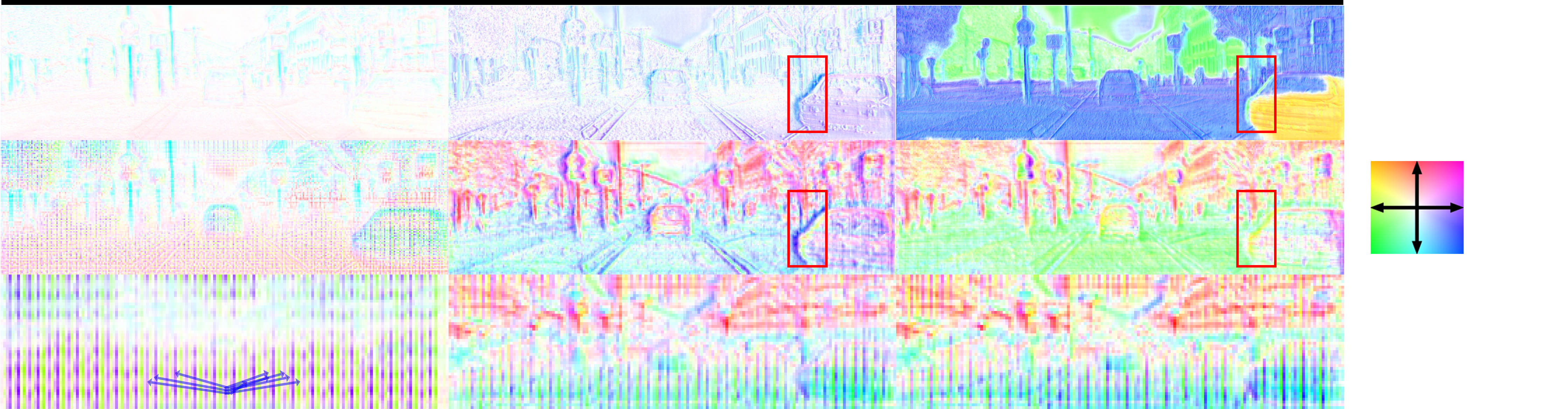}
  \leftline{ \qquad\qquad\qquad\quad\  $\Delta_{F}$ \qquad\qquad\qquad\qquad\quad $\Delta_{C1}$\qquad\qquad\qquad\qquad\ \  $\Delta_{C2}$}
  \centerline{(b)}
  \caption{(a) Visualization of the raw and distilled depth maps predicted with the SDFA-Net on KITTI.
           (b) Visualization of the corresponding offset maps learnt in SDFA modules.
           From the top to bottom, the offset maps are learnt in SDFA-1, SDFA-2, and SDFA-3.
           The color coding is visualized on the right part, where the hue and saturation denote the norm and direction of the offset vector respectively.
           The blue arrows in the bottom left offset map show the offset vectors in a $3 \times 3$ patch.
  }
  \label{fig:visual}
\end{figure}

From Lines 1-5 of Table~\ref{tab:abla}(a), it is noted that using the modified version ( `Swin$^\dagger$') of Swin transformer performs better than its original version for depth prediction, indicating that the high-resolution features with rich spatial details are important for predicting depths.
By replacing different decoder blocks with the proposed SDFA modules, the performances of the models are improved in most cases.
It demonstrates that the SDFA moudule could aggregate features more effectively for depth prediction. 
As seen from Lines 6-7 of Table~\ref{tab:abla}(a), although `Swin$^\dagger$+OA' achieves better quantitative performance compared to `Swin$^\dagger$+Raw', our full model performs best under all the metrics, which indicates that the SDFA module is benefited from the offset maps learnt under the self-distillation.

Additionally, we visualize the depth maps predicted with the models denoted by `Swin$^\dagger$+Raw', `Swin$^\dagger$+OA', and our SDFA-Net in Figure~\ref{fig:head}.
Although the offset-based aggregation improves the quantitative results of depth prediction, `Swin$^\dagger$+OA' is prone to predict inaccurate depths on occlusion regions.
SDFA-Net does not only alleviate the inaccurate predictions in these regions but also further boost the performance of depth estimation.
It demonstrates that the SDFA module maintains the contextual consistency between the learnt features for more accurate depth prediction.
Please see the supplemental material for more detailed ablation studies.

To further understand the effect of the SDFA module, in Figure~\ref{fig:visual}, we visualize the predicted depth results of SDFA-Net, while the feature offset maps learnt by each SDFA module are visualized with the color coding.
Moreover, the average norms of the offset vectors in these modules are shown in Table~\ref{tab:abla}(b), which are calculated on the raw Eigen test set~\cite{Geiger2012We}.
As seen from the first column in Table~\ref{tab:abla}(b) and Figure~\ref{fig:visual}(b), the offsets in $\Delta_F$ reflect the scene information, and the norms of these offset vectors are relatively long.
These offsets are used to refine the up-sampled decoded features with the captured information.
Since the low-resolution features contain more global information and fewer scene details, the directions of the offset vectors learnt for these features (e.g.,  $\Delta_F^3$ in bottom left of Figure~\ref{fig:visual}(b)) vary significantly.
Specifically, considering that the features in an arbitrary $3 \times 3$ patch are processed by a convolutional kernel, the offsets in the patch could provide the adaptive non-local information for the process as shown with blue arrows in Figure~\ref{fig:visual}(b).
As shown in the last two columns in Table~\ref{tab:abla}(b) and Figure~\ref{fig:visual}(b), the norms of the offsets in $\Delta_{C1}$ and $\Delta_{C2}$ are significantly shorter than that in $\Delta_F$.
Considering that the encoded features maintain reliable spatial details, these offsets are used to refine the features in little local areas.

Moreover, it is noted that there are obvious differences between the $\Delta_{C1}$ and $\Delta_{C2}$, which are learnt under the self-supervised and self-distillation manners respectively.
Meanwhile, the depth map obtained with the features aggregated by $\Delta_F$ and $\Delta_{C2}$ are more accurate than that obtained with the features aggregated by $\Delta_F$ and $\Delta_{C1}$, especially on occlusion regions (marked by red boxes in Figure~\ref{fig:visual}(a)).
These visualization results indicate that the learnable offset maps are helpful for adaptively aggregating the contextual information, while the self-distillation is helpful for further maintaining the contextual consistency between the learnt feature, resulting in more accurate depth predictions.

\section{Conclusion}
In this paper, we propose the Self-Distilled Feature Aggregation (SDFA) module to adaptively aggregate the multi-scale features with the learnable offset maps.
Based on SDFA, we propose the SDFA-Net for self-supervised monocular depth estimation, which is trained with the proposed self-distilled training strategy.
Experimental results demonstrate the effectiveness of the SDFA-Net.
In the future, we will investigate how to aggregate the features more effectively for further improving the accuracy of self-supervised monocular depth estimation.

\spnewtheorem*{acks}{Acknowledgements}{\bf}{}
\begin{acks}
This work was supported by the National Natural Science Foundation of China (Grant Nos. U1805264 and 61991423), the Strategic Priority Research Program of the Chinese Academy of Sciences (Grant No. XDB32050100), the Beijing Municipal Science and Technology Project (Grant No. Z211100011021004).
\end{acks}

\clearpage
%
%
\bibliographystyle{splncs04}
\bibliography{egbib}
\clearpage

\appendix
\renewcommand{\maketitle}{
 \
 \null
 \begin{center}
  {\Huge Supplementary Material \par}
 \end{center}%
 \par}
\maketitle

\setcounter{figure}{0}
\setcounter{table}{0}

\renewcommand{\theequation}{A\arabic{equation}}
\renewcommand{\thetable}{A\arabic{table}}
\renewcommand{\thefigure}{A\arabic{figure}}

\section{Datasets and Metircs}
The two datasets used in this work are introduced in detail as follows: 
\begin{itemize}
\item KITTI~\cite{Geiger2012We} contains the rectified stereo image pairs captured from a driving car.
We use the Eigen split~\cite{Eigen2014Depth} to train and evaluate the proposed network (called SDFA-Net), which consists of 22600 stereo image pairs for training and 697 images for testing.
Additionally, we also evaluate SDFA-Net on the improved Eigen test set, which consists of 652 images and adopts the high-quality ground-truth depth maps generated with the method in~\cite{Uhrig2017Sparsity}.
The images are resized into the resolution of $1280 \times 384$ at both the training and inference stages, while we assume that the intrinsics of all the images are identical.  

\item Cityscapes~\cite{Cordts2016The} contains the stereo pairs of urban driving scenes, and we take 22972 stereo pairs from it for jointly training SDFA-Net.
When SDFA-Net is trained on both the KITTI and Cityscapes datasets, we crop and resize the images from Cityscapes into the resolution of $1280 \times 384$.
Considering that the baseline length in Cityscapes is different from that in KITTI, we scale the predicted disparities on Cityscapes by the rough ratio of the baseline lengths in the two datasets.
\end{itemize}

For the evaluation on both the raw and improved KITTI Eigen test set~\cite{Eigen2014Depth}, we use the center crop proposed in~\cite{Garg2016Unsupervised} and the standard depth cap of 80m.
The following metrics are used:
\begin{itemize}
    \item Abs Rel: $\frac{1}{N}\sum_i{\frac{\left| \hat{D}_i - D^{gt}_i \right|}{D^{gt}_i}}\quad,$
    
    \item Sq Rel: $\frac{1}{N}\sum_i{\frac{\left| \hat{D}_i - D^{gt}_i \right|^2}{D^{gt}_i}}\quad,$
    
    \item RMSE: $\sqrt{\frac{1}{N}\sum_i{\left| \hat{D}_i - D^{gt}_i \right|^2}}\quad,$
    \item logRMSE: $\sqrt{\frac{1}{N}\sum_i{\left| \log\left(\hat{D}_i\right) - \log\left(D^{gt}_i\right) \right|^2}}\quad,$
    \item Threshold (A$j$):  $ \% \quad s.t. \quad \max{\left( \frac{\hat{D}_i}{D^{gt}_i}, \frac{D^{gt}_i}{\hat{D}_i} \right)}< a^{j}\quad,$
\end{itemize}
where $\{\hat{D}_i, D^{gt}_i\}$ are the predicted depth and the ground-truth depth at pixel $i$, and $N$ denotes the total number of the pixels with the ground truth.
In practice, we use $a^{j} = 1.25, 1.25^{2}, 1.25^{3}$, which are denoted as A1, A2, and A3 in all the tables.

\section{Ablation studies on the self-distilled training strategy}
We conduct more ablation studies on the KITTI dataset~\cite{Geiger2012We} for verifying the effectiveness of the proposed self-distilled training strategy.
We firstly train a model that uses the Offset-based Aggregation (OA) modules in the decoder with a straightforward Self-Distilled training strategy (`Swin$^\dagger$+OA (SD)').
Specifically, since the OA module does not have the `Distilled data path' (described in Section 3.2), this model is trained under the self-distillation manner by using the `Raw data path' twice in the two steps of each training iteation (described in Section 3.3).
The corresponding results are shown in Lines 1-3 of Table~\ref{tab:supp}.
The results predicted by the `Swin$^\dagger$+OA' (without the self-distillation) and the full model `SDFA-Net' are also reported for comparison.
It can be seen that there are only slight improvements under four metrics when the straightforward self-distilled strategy is used. 
Our full model performs best under all the metrics, which indicates that the proposed self-distilled training strategy with the two data paths is more helpful for predicting accurate depths.

\begin{figure}[h]
  \centering
  \includegraphics[width=12cm]{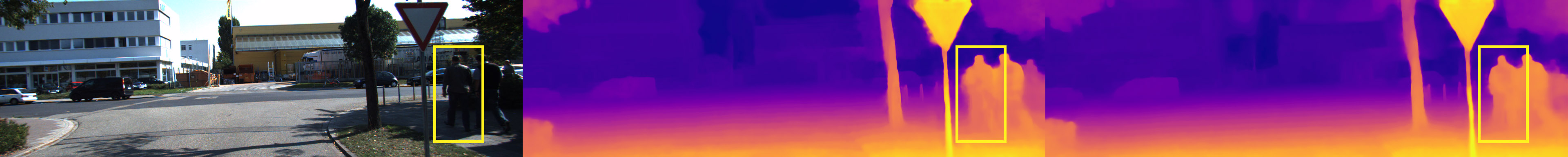}
  \includegraphics[width=12cm]{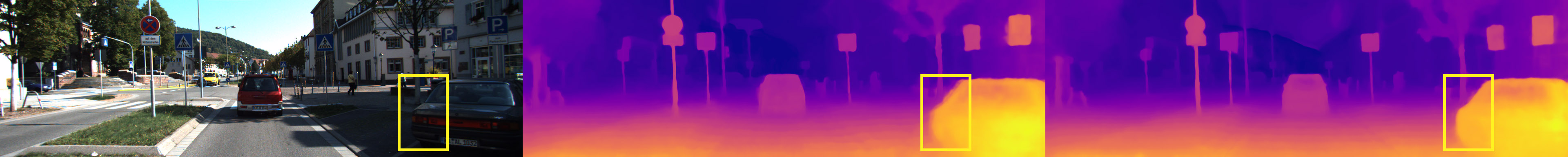}
  \scriptsize
  \leftline{\qquad\qquad\ \  Input Images \qquad\qquad\qquad\quad SDFA-Net w/o. Flip \qquad\qquad\qquad\quad SDFA-Net}
  \caption{Visualization results of the SDFA-Net trained without/with the feature flipping strategy on KITTI.
  }
  \label{fig:flip}
\end{figure}

To verify the effectiveness of the principle masks and feature flipping strategy defined under the self-distilled training strategy, we conduct further ablation studies by omitting the visible principle mask (`w/o. $M^l_v$'), the photometric principle mask (`w/o. $M^l_p$'), both of the masks (`w/o. Masks'), and the feature flipping (`w/p. Flip'), respectively.
The results shown in Lines 4-7 of Table~\ref{tab:supp} demonstrate that both of the masks could improve the accuracy, and the principle mask has a stronger influence than the visible mask.
It also can be seen that the flipped features are more helpful for improving the accuracy.
Since the weights of `Raw \& distilled data path' (described in Section 3.2) in SDFA are shared at the self-supervised and self-distilled forward propagation steps, they are trained by minimizing both the image synthesis loss and self-distilled loss.
Therefore, the distilled depths predicted in the Self-distilled Forward Propagation are still suffer from occlusions to some extent.
Specifically, the model trained with left images of stereo pairs under a self-supervised manner always predicts inaccurate depths on the left side of objects, whether the input features are flipped or not.
The feature flipping strategy could alleviate the occlusion problem because the inaccurate depths would occur on the right side of objects in the distilled depth maps when the input features are flipped, which are not on the real occluded regions and could be corrected by the reliable pseudo depths.
The visualization results of the depths predicted by the model trained with and without feature flipping shown in Figure~\ref{fig:flip} also illustrate that our full model predicts sharper and more accurate depths on occluded regions compared to `SDFA-Net w/o. Flip'.
It demonstrates the effectiveness of the feature flipping strategy.

\section{Ablation study on the backbone}
To further explore the effectiveness of the proposed SDFA module on different backbones, we train a new baseline model comprises a ResNet18~\cite{he2016deep} as the encoder and a raw decoder proposed in~\cite{Godard2019Digging} (Res18+Raw).
Then we use the ResNet18~\cite{he2016deep} to replace the modified Swin-transformer~\cite{liu2021swin} in SDFA-Net (SDFA-Net (Res18)).
The corresponding results are reported in Lines 8-9 of Table~\ref{tab:supp}.
It can be seen that SDFA also could improve the performance of the ResNet18-based model.

\begin{table}[h]
  \centering
  \caption{Additional quantitative results on the raw KITTI Eigen test set in the ablation study.}
  \scriptsize
  \renewcommand\tabcolsep{1.2pt}
  \begin{tabular}{|l|cccc|ccc|}
     \hline
     \multicolumn{1}{|c|}{Method}         & Abs Rel$\downarrow$ & Sq Rel$\downarrow$ & RMSE$\downarrow$  & logRMSE$\downarrow$ & A1$\uparrow$    & A2$\uparrow$    & A3$\uparrow$    \\\hline
     Swin$^\dagger$+OA     & 0.091   & 0.554  & 4.082 & 0.174   & 0.899 & 0.967 & 0.984 \\
     Swin$^\dagger$+OA (SD) & 0.092   & 0.549  & 4.003 & 0.174   & 0.900 & 0.967 & 0.985 \\
     SDFA-Net   & 0.090   & 0.538  & 3.896 & 0.169   & 0.906 & 0.969 & 0.985 \\\hline
     SDFA-Net w/o. $M^l_v$ & 0.090   & 0.544  & 3.928 & 0.169  & 0.905 & 0.969 & 0.985 \\
     SDFA-Net w/o. $M^l_p$ & 0.096   & 0.566  & 4.013 & 0.173  & 0.902 & 0.968 & 0.985\\
     SDFA-Net w/o. Masks   & 0.095   & 0.583  & 4.039 & 0.175  & 0.899 & 0.967 & 0.985  \\
     SDFA-Net w/o. Flip    & 0.097   & 0.583  & 4.078 & 0.177   & 0.897 & 0.966 & 0.984 \\\hline
     Res18+Raw             & 0.102   & 0.644  & 4.341 & 0.187   & 0.880 & 0.960 & 0.982  \\
     SDFA-Net (Res18)      & 0.101   & 0.636  & 4.226 & 0.180   & 0.891 & 0.964 & 0.984 \\\hline
  \end{tabular}
  \label{tab:supp}
\end{table}

\end{document}